\documentclass[runningheads]{llncs}
\usepackage[T1]{fontenc}
\usepackage{graphicx}
\usepackage{cite}
\usepackage{amsmath,amssymb,amsfonts}
\usepackage{algpseudocode}
\usepackage{algorithm}

\usepackage{graphicx}
\usepackage{textcomp}
\usepackage{booktabs}
\usepackage{xcolor}
\usepackage{float}
\usepackage{url}
\def\BibTeX{{\rm B\kern-.05em{\sc i\kern-.025em b}\kern-.08em
    T\kern-.1667em\lower.7ex\hbox{E}\kern-.125emX}}
\usepackage{color}

\usepackage{subcaption}
\usepackage{multirow}

\begin{document}
\title{Curriculum-Based Adversarial Heterogeneous Agent Reinforcement Learning for Autonomous Quad-Copter Landing in Maritime Settings}
\titlerunning{HARL-AC for Autonomous Maritime Quad-Copter Landing}
%
\author{Allan Minh-Tam Nguyen \orcidID{0009-0001-7687-6741} \and Sree Showrya Kotala \orcidID{0009-0001-0929-4514} \and
Stefan Banioi-Crijman \orcidID{0009-0004-5082-606X} \and
Kurt Driessens \orcidID{0000-0001-7871-2495} \and
Rico Möckel \orcidID{0000-0001-5497-3754}
}
\authorrunning{A. Nguyen et al.}
%
\institute{Department of Advanced Computing Sciences, Maastricht University, Minderbroedersberg 4-6, 6211 LK Maastricht, Netherlands
\email{rico.mockel@maastrichtuniversity.nl} 
}

\maketitle

\begin{abstract}
Recovering unmanned aerial vehicles (UAVs) in maritime environments is challenging due to wind turbulence and ship-deck motion, making it a valuable test case for alternative control and learning approaches as conventional landing approaches often become unreliable. 
We study simulated mid-air capture of quadrotor UAVs by a ship-mounted robotic arm, learning robust cooperative control policies with Heterogeneous-Agent Proximal Policy Optimization (HAPPO) Reinforcement Learning. 
We train with HAPPO using a curriculum and an adversarial wind agent (HARL-AC) in NVIDIA Isaac Lab, and compare the obtained control policies against those generated through curriculum-based domain randomization and a benchmark trained on a single sea state.
In-distribution evaluation on sea states $0/4/5$ shows comparable success for HARL-AC and domain randomization of up to $97.5\%$. 
On out-of-distribution sea states $7/8/10$, HARL-AC generalizes better, achieving up to $16\%$ higher median success rate at sea state 10, and substantially lower crash rates of up to $14\%$ compared to the domain randomization policy. 
Furthermore, we show that the adversarially trained policy shows more cautious behavior, slightly increasing timeouts by $<3\%$, but yields safer recovery behavior in severe, unseen conditions.
 
\end{abstract}


\section{Introduction}
The safe landing of an Unmanned Aerial Vehicle (UAV) is critical to achieve fully autonomous operations. 
Current autonomous landing approaches circumvent the dangers from interactions of the turbulent rotor wake and landing surface by a hovering at a low altitude then cutting power to the motors, effectively dropping the drone onto the landing platform from a `safe' height.  
This approach works for stable, large landing platforms when external perturbations such as wind are not strong enough to push the falling drone off the platform. 
In a less forgiving environment, for example when landing on the deck of a small ship or in stronger wind conditions, conventional landing approaches can result in damage to the equipment \cite{Keipour2022VisualServoingUAVLanding, Tian2024SeaAirHeterogeneousSystem}. 
This paper explores an alternative approach to retrieve flying drones. Using a robotic arm to catch the airborne drone, a landing platform can be avoided and small movements of the boat can be compensated. Also, avoiding the use of a landing platform removes the generation of turbulent wakes. These benefits come at the cost of requiring cooperated multi agent control, coordinating the behavior of the drone and the robotic arm. 
To address this challenge, we compare Reinforcement Learning (RL) strategies in which the UAV and robot arm control policies are derived using a Curriculum-Based Domain Randomization (DR) approach against a Curriculum-Based Adversarial Heterogeneous Agent Reinforcement Learning (HARL-AC) approach. 
The prevailing wind and gusts, in addition to the reactive movement of the ship deck introduce stochastic disturbances into the environment and provides a meaningful test medium to compare between the two derived control policies.

Recent work by  Huang et al. \cite{Huang2026SafeUAVWind} presents a Deep Reinforcement Learning controller that attempts to address the issue of windy conditions for UAVs by optimizing the drone's response to aerodynamic effects. 
However, this paradigm relies on an assumption of strong and responsive thrust capabilities of the UAV in highly windy and dynamic conditions. 
To rely less on this assumption, this work adopts an alternative of assisting the landing procedure via mid-air capture.
While the mechanical viability of catching a drone mid-air has been previously demonstrated using soft robotic effectors and visual tracking in controlled environments \cite{Fedoseev2021DroneTrap}, this work begins to explore the potential for an autonomous controller in a more unpredictable maritime environments.

Coordinating the heterogeneous interception requires a specialized Multi-Agent Reinforcement Learning framework. 
Standard algorithms like MAPPO \cite{Yu2022PPOMultiAgentGames} rely on parameter sharing across a policy. 
A MAPPO formulation works well for homogeneous multi-agents settings, however for heterogeneously formulated agents, it exposes itself to value collapse where the learning agent tries to overgeneralize a single policy \cite{Zhong2024HeterogeneousAgentRL}. 
Given the degree of heterogeneity of the robot arm and UAV formulations, this work employs Heterogeneous-Agent Proximal Policy Optimization (HAPPO) \cite{PKUMARL2024HARL}, which provides monotonic improvement guarantees and evolves independent policies.

While stochastic DR is the standard baseline for robust sim-to-real transfer \cite{Tobin2017DomainRandomization}, it frequently fails to expose agents to worst-case boundary conditions, flattening the learning gradient \cite{Pinto2017RobustAdversarialRL, Dennis2020UnsupervisedEnvironmentDesign}. 
To enforce genuine disturbance rejection, this work builds upon previous work \cite{BanioiDronesSoftware} in IsaacLab\cite{Mittal2025IsaacLab} and implements a zero-sum adversary. 
Bound by a dynamic curriculum, the adversarial wind agent actively disrupts the successful capture of the drone using the robot arm end effector by injecting generated disturbances  simulating windy gusts and corresponding wave conditions in sea states up to 5, further elaboration for these is included in Section \ref{sec:adversarial_action_space}.
This work explores the differences during the learning process and deployment characteristics between the two heterogeneous agent controllers.

\section{Related work}
Under the centralized training decentralized execution paradigm, MAPPO~\cite{Yu2022PPOMultiAgentGames} is a standard baseline but relies on parameter sharing. 
While effective for homogeneous multi-agent systems, policy parameter sharing forces abstractions or generalizations of structurally distinct agents, resulting in a flattened policy representation instead of training a policy that accounts for the heterogeneity of agents. 
As a result, the learning process is susceptible to training instability and lacks convergence guaranties \cite{Zhong2024HeterogeneousAgentRL}. 
To resolve this, HAPPO~\cite{Zhong2024HeterogeneousAgentRL} utilizes a sequential, per-agent update scheme. 
This Heterogeneous Agents Reinforcement Learning (HARL) approach guarantees monotonic improvement for individual agents without requiring shared network weights \cite{Zhong2024HeterogeneousAgentRL}.

To bridge the sim-to-real gap, Robust Adversary Reinforcement Learning \cite{Pinto2017RobustAdversarialRL} frames policy learning as a zero-sum game where an adversary applies physical disturbances to optimally destabilize the system. 
However, scaling this to multi-agent teams using a shared critic causing severe credit assignment failure as the advantage function vanishes due to the critic estimate collapsing to the average even if one team wins.~\cite{Peterson2025IsaacLabHeterogeneousAdversarial}. 
This then results in a degraded PPO loss function and no meaningful updates can then be made to either teams policy.
The Heterogeneous Multi-Agent Adversarial Reinforcement Learning (HARL-A) framework proposed by Peterson et al. \cite{Peterson2025IsaacLabHeterogeneousAdversarial} resolves this by integrating team-separated critics into HAPPO’s sequential update scheme, enabling stable adversarial gradients across heterogeneous teams.

Because extreme adversarial conditions often prevent initial policy convergence, curriculum learning is required to progressively scale task difficulty. 
Xiao et al. \cite{xiao_collaborative_2025} showcase a cooperative approach with curriculum learning for visual search with a swarm of drones.
More recently Yan et al. \cite{yan_curriculum-guided_2026} apply the curriculum-based RL paradigm where a base station facilitates communication services between the UAVs and fuses sensor data from multiple UAVs so the base station and UAVs can jointly track a target.
In cooperative HARL environments, introducing pre-trained teammates in a cooperative setting accelerates coordination. 
Bhati et al.~\cite{Bhati2023CurriculumCooperationMARL} find that when comparing a curriculum of teammates with just a single pre-trained teammate helping the agent, a medium-skilled pre-trained teammate strikes the optimal balance between the overall team reward and individual learning for the agent. 

Consequently, training robust heterogeneous policies requires synthesizing both approaches: utilizing stage-based progression and individual pre-training to organically scale adversarial intensity.

Some work thus far has addressed the gap where both curriculum and adversarial learning have been explored in conjunction. 
Xu et al. \cite{xu_heterogeneous_2025} present heterogeneous adversarial play in which a student-teacher framework is used to develop both agents in an adversarial zero-sum manner. 
The teacher-student methodology is applied to various frameworks, including Crafter \cite{pmlr-v70-andreas17a},  a multi-task RL environment inspired by Minecraft that, similar to the chosen setting of this work, introduces stochastic elements and open-world elements.
A slightly earlier work by Kim et al. \cite{kim_learning_2025} implements adversarial-cooperative HARL for two robot arms that are performing a throwing catching task. 
Although the physical structures and thus representation of the agents are homogeneous, the policies trained for the throwing arm and catching arm are necessarily distinct. 

Extending the related work, this work develops insight on generating robust cooperative control through the combination of curricula and adversarial multi-agent reinforcement learning in the instance of a cooperative team of heterogeneous agents competing against an adversary.

\section{Methods}

\begin{figure*}[ht]
    \centering
    \includegraphics[width=\linewidth,height = 8cm]{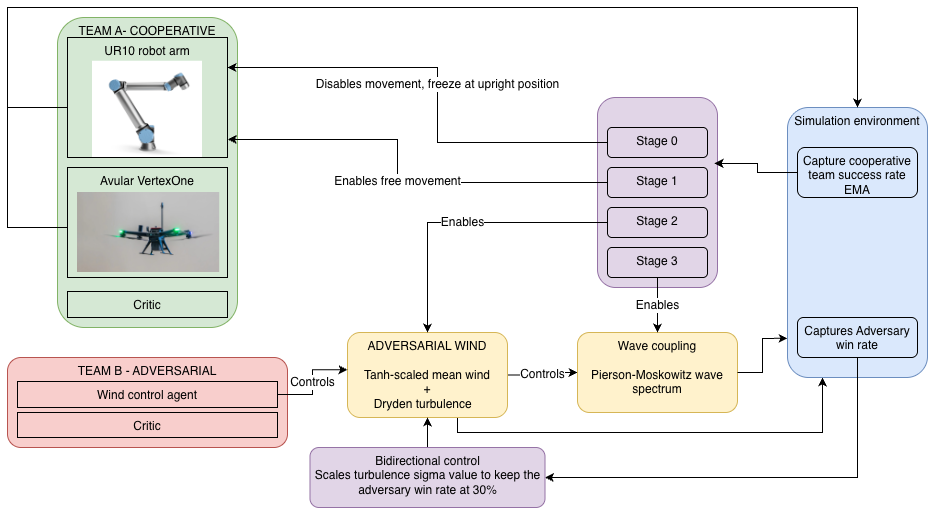} 
    \caption{Flowchart outlining the curriculum components used in combination with adversarial training for the cooperative drone and arm against the wind adversary and environment. The curriculum introduces increasing complexity as stages are progressed in simulation.}
    \label{fig:visual_map_curriculum}
\end{figure*}

An outline of the HARL-AC training regimen is provided in Fig. \ref{fig:visual_map_curriculum}. The individual components are detailed below.

\subsection{Multi-agent formulation}
The task of safe rendezvous between the end effector of the robot arm and quadcopter is formulated as a heterogeneous multi-agent task with three agents organized into two teams.
Team A, the cooperative team, includes a simulated UR10 robot arm and an Avular Vertex One quadcopter UAV, with the joint goal of successful connection between the robot arm end effector and drone body. 
A successful landing is assumed when a fixed point under the center of the drone body can remain within a $0.5m$ bubble above the UR10 end effector (EE) for $3$ seconds while maintaining a speed less than $3m/s$ for this duration. 
This was chosen to simplify the `catch' procedure and provide results agnostic to the catch mechanism. 
Team B, the adversarial team, consists of the wind component aimed at preventing the drone and arm from succeeding in their task. 

The agents' observations encompass the following state variables (total number of observed state dimensions per agent given in brackets): 
For the drone (23-Dimensional State Space (D)): drone position, quaternion orientation, linear and angular velocities, previous thrust/direction action, the location of the robot arm end-effector, and wind forces. 
For the UR10 arm (37-D): joint positions/velocities, end-effector pose and velocities, previous 6-joint control actions, drone position, and wind.
For the wind adversary (19-D): relative drone-to-goal distance, drone state (velocity and quaternion orientation), end-effector position, and current wind momentum.

The action spaces for the agents include the following: for the UAV (4-D): the thrust and 3 axis angular rotation values (pitch, yaw, roll), for the UR10 (6-D): the joint position targets and for the wind adversary (3-D): the 3 axis wind force that is superimposed as a disturbance onto the final positions of Team A. 

Training uses Heterogeneous-Agent Proximal Policy Optimization (HAPPO) \cite{PKUMARL2024HARL} with team-separated critics via the HARL-A framework proposed by Peterson et al.~\cite{Peterson2025IsaacLabHeterogeneousAdversarial}. 
Unlike simultaneous update algorithms, HAPPO updates policies sequentially, conditioning each agent's update on the new policies of preceding agents. 
This provides theoretical guarantees of monotonic improvement without parameter sharing \cite{Zhong2024HeterogeneousAgentRL}, preventing training collapse caused by a shared policy between agents in highly asymmetric tasks. 
Furthermore, the critics remain strictly separated: Team A's critic evaluates a centralized 60-dimensional state (concatenating both cooperative agents' observations), while Team B's critic independently processes the adversary's 19-dimensional state \cite{Peterson2025IsaacLabHeterogeneousAdversarial}.

\subsection{Adversarial model}
\subsubsection{Adversarial Action Space and Target Mapping:}
\label{sec:adversarial_action_space}

Within the multi-agent reinforcement learning framework, the wind adversary functions as an independent policy operating within a continuous, three-dimensional action space. These dimensions directly correspond to the scaled target wind vector mapped to the global Cartesian coordinate system ($X$, $Y$, and $Z$ axes). 

To guarantee numerical stability and to enforce the physical constraints defined by the active curriculum stage, the raw actions generated by the adversarial policy, denoted as $\vec{a}_{\text{adv}}$, undergo a sequence of transformations. 
First, the network outputs are bounded to the interval $[-1.0, 1.0]$. 
Subsequently, a hyperbolic tangent activation function is applied, and the result is scaled by $V_{\text{max}}$, representing the maximum permissible wind speed for the current training stage. 
This formulation yields the bounded target wind velocity vector, $\vec{w}_{\text{target}}$, defined as: $\vec{w}_{\text{target}} = V_{\text{max}} \cdot \tanh(\vec{a}_{\text{adv}})$.

\subsubsection{Dryden turbulence filter:}
To simulate realistic aerodynamic disturbances, the environment incorporates a dynamic wind model combining adversary-controlled wind with stochastic Dryden turbulence\cite{MathWorksDrydenWindTurbulenceModel}. 

At each simulation step $\Delta t$, the wind vector is updated in two primary steps: 

\noindent\textbf{Step 1: Adversarial Mean Wind}
The adversary's continuous actions are scaled via a hyperbolic tangent function to define a target wind velocity. 
To prevent unrealistic, instantaneous shifts, this target vector passes through a first-order low-pass filter. The smoothing factor $\alpha_{\text{ctrl}} = \exp(-\Delta t / \tau)$ is determined by the time constant $\tau = \frac{L}{V}$, where $L$ is the turbulence scale length and $V$ is the nominal airspeed. 
This exponential moving average ensures that the mean wind vector $\vec{\mu}_{\text{wind}}$ transitions smoothly over time.

\noindent\textbf{Step 2: Stochastic Dryden Turbulence}
To simulate atmospheric variations, stochastic turbulence $\vec{x}_{\text{turb}}$ is superimposed onto the mean wind. 
The turbulence is modeled as a discrete-time first-order Markov process approximating the continuous Dryden model\cite{MathWorksDrydenWindTurbulenceModel}. 
The state is driven by standard Gaussian white noise $\mathcal{N}(0, I)$, scaled by the standard deviation $\sigma$ and a spatial frequency factor $\alpha_{\text{dryden}} = \frac{V \Delta t}{L}$. 
The turbulence update follows the stochastic difference equation: $\vec{x}_{t} = (1 - \alpha_{\text{dryden}}) \vec{x}_{t-1} + \sigma \sqrt{2 \alpha_{\text{dryden}}} \mathcal{N}(0, I)$.

Finally, the smoothed adversarial wind and the instantaneous turbulence are summed. 
To enforce the physical limits of the training domain, the combined velocity vector is clamped to a maximum magnitude $V_{\text{max}}$ before being applied as an external force to the UAV (see Appendix A for pseudocode).
Tweaking the $V_{max}$ and $\sigma$ allows for various sea states to be represented during training and evaluation.

\subsubsection{Wind-Wave Coupling:}
 Unlike conventional adversarial RL where the adversary controls a single disturbance channel, this adversary's actions are coupled to ocean dynamics that are introduced in the last stage of the curriculum. 
 The instantaneous wind magnitude drives a slow exponential filter on the sea state:
\begin{equation*}
U_{t+1} = \alpha_{\text{sea}} U_t + (1 - \alpha_{\text{sea}}) \|w_t\|, \quad \alpha_{\text{sea}} = e^{-\Delta t / \tau_{\text{sea}}}
\label{eq:sea_state}
\end{equation*}
with $\tau_{\text{sea}} = 0.5$~s. The filtered sea state $U_t$ parameterise a Pierson--Moskowitz wave spectrum~\cite{Fossen2011HandbookMarineCraft}:
\begin{equation*}
S(\omega) = \frac{\alpha_{\text{PM}} g^2}{\omega^5} \exp\!\left(-\beta_{\text{PM}} \left(\frac{g}{U \omega}\right)^4\right)
\label{eq:pm_spectrum}
\end{equation*}
with $\alpha_{\text{PM}} = 0.0081$, $\beta_{\text{PM}} = 0.74$, sampled at 8 frequency components $\omega \in [0.4, 3.0]$~rad/s. 
Spectral amplitudes $A_i = \sqrt{2 S(\omega_i)\, \Delta\omega}$ are realized as platform motion across five degrees of freedom (surge, sway, heave, roll, pitch) via a sum-of-sines with random per-environment phases:
\begin{equation*}
\delta x(t) = \sum_{i=1}^{8} c_x A_i \sin(\omega_i t + \phi_i^{x})
\end{equation*}
with translational response coefficients $c = [0.2, 0.1, 1.0]$ for (surge, sway, heave) and rotational amplitudes additionally scaled by wave number $k_i = \omega_i^2 / g$ with rotational coefficients $c_{\text{rot}} = [0.5, 1.0]$ for (roll, pitch). 
Heave and pitch coefficients are gated to zero in early curriculum stages (Table~\ref{tab:curriculum}). 
This coupling produces a compound disturbance: the adversary's actions perturb the UAV aerodynamically and induce platform motion that displaces the robot arm's base.
The displacement components of surge, heave and sway were empirically determined to create sufficiently difficult landing conditions such that the standard motor cutoff approach would be unreliable and risk damage to the UAV. 

\subsection{Reward structure}
Following HARL-A's reward structure \cite{Peterson2025IsaacLabHeterogeneousAdversarial}, each team calculates their reward independently. 
Team A gets rewards based on their shared cooperation, while team B's reward is formulated by the zero-sum component against team A plus adversary-specific shaping terms. 
Active reward terms are listed in Appendix D.

\subsection{Curriculum learning}
The system is trained in a 4-stage curriculum that progressively introduces adversarial variables into the environment as the stages increase. 
Stage 0 is reserved for pre-training by freezing the arm upright and letting the drone learn by itself first. 
The adversarial pressure is rolled out starting from stage 2 (Table \ref{tab:curriculum})
\begin{table}[!t]
\centering
\caption{Curriculum stage components with increasing difficulty as stages progress. Sigma corresponds with the wind disturbance and the values are chosen to reflect sea states 4 and 5. Tight limits correspond to $0.3m-0.8m$ around the center of the environment, normal limits correspond to $0m- 1.7m$. Ocean effects comprise of the heave, pitch. yaw, roll, surge and sway.}
\label{tab:curriculum}
\small
\setlength{\tabcolsep}{4pt}
\begin{tabular}{c@{\hspace{6pt}}c@{\hspace{6pt}}c@{\hspace{6pt}}c@{\hspace{6pt}}c}
\toprule
\textbf{Stage} & \textbf{Sigma value} & \textbf{Ocean Effects} & \textbf{Spawn} & \textbf{Bounds} \\
\midrule
0 & 0 & $\times$ & tight & tight \\
1 & 0 & $\times$ & normal & normal \\
2 & 4 & $\times$ & normal & normal \\
3 & 7 & $\checkmark$ & normal & normal \\
\bottomrule
\end{tabular}
\end{table}

\textbf{Stage progression:} Stage progression is controlled by an exponential moving average (EMA) of the success rate, updated in batches of $N_{\text{batch}} = 512$ episodes:
\begin{equation*}
\bar{r}_{t+1} = (1 - \beta) \bar{r}_t + \beta r_t, \quad \beta = \begin{cases} 0.02 & r_t > \bar{r}_t \\ 0.10 & r_t \leq \bar{r}_t \end{cases}
\label{eq:ema}
\end{equation*}
The asymmetric smoothing reflects a conservative design philosophy: stage advancement requires sustained improvement, reflected by a slow EMA rise, to ensure the policies have genuinely converged before facing harder conditions, while stage retreat is rapid, reflected by a fast EMA decline, to escape regimes where the cooperative pair has lost stability. 
This asymmetry mitigates the oscillation between stages. 
The stage advances when $\bar{r}_t > 0.6$ and retreats when $\bar{r}_t < 0.3$. 
After each transition, a 5-batch warm-up suppresses EMA updates while the policies adapt.

\subsection{Bidirectional adversarial curriculum}
Unconstrained mini-max adversaries often cause training collapse by diverging to trivial extremes, i.e., physically impossible winds or no wind at all, flattening the learning gradient \cite{Pinto2017RobustAdversarialRL, Dennis2020UnsupervisedEnvironmentDesign}. 
To maintain an optimal challenge and prevent this collapse, the environment employs a bidirectional curriculum to dynamically scale high-frequency turbulence.

The curriculum anchors the adversary's win rate to a target threshold $T_{\text{adv}} = 0.30$ (a 30\% adversarial win rate), with an acceptable tolerance band $\delta = 0.10$. 
At the end of each batch, the Dryden turbulence standard deviation $\sigma$ is multiplicatively updated:
\begin{equation*}
    \sigma_{t+1} = 
    \begin{cases} 
      \min(\sigma_t \times (1 + \eta), \sigma_{\text{max}}), & \text{if } R_{\text{adv}} < (T_{\text{adv}} - \delta) \\
      \max(\sigma_t \times (1 - \eta), \sigma_{\text{min}}), & \text{if } R_{\text{adv}} > (T_{\text{adv}} + \delta) \\
      \sigma_t, & \text{otherwise}
    \end{cases}
    \label{eq:bidir_sigma}
\end{equation*}
where $\eta = 0.05$ represents the fractional step size. 
The turbulence intensity is strictly clamped within predefined, stage-specific boundaries $[\sigma_{\text{min}}, \sigma_{\text{max}}]$ to ensure that aerodynamic perturbations remain physically plausible for the simulated environment.

Ultimately, this reformulates training into a constrained mini-max game. 
Bound to a 30\% win rate, the adversary cannot rely on brute-force magnitude to guarantee crashes. 
Instead, it must continuously hunt for the threshold of solvability, organically exploring a much wider and more sophisticated variety of wind conditions to exploit the drone's vulnerabilities.

\section{Training and Evaluation}

\subsection{Ablation Study: Stochastic Wind with Curriculum (Pure Dryden)}
To evaluate whether a learned adversarial policy yields better results than standard domain randomization approaches for HARL curriculum learning, a stochastic wind baseline was implemented. 
In this configuration, the adversarial component of the wind agent is entirely removed from the simulated curriculum. 
Instead, the wind's action is fully determined by a white noise generator passed through the Dryden turbulence filter.
Without the bidirectional curriculum to auto-scale the difficulty, the turbulence intensity ($\sigma$) in this baseline is statically parametrized based on the macroscopic curriculum stage, defined as $\sigma = 4$ in Stage 2 and $\sigma = 7$ in Stage 3. 

The total environmental wind energy was bifurcated into two independent, superimposed stochastic processes:

\textbf{Slow, Rolling Thermals:} Modeled using a doubled spatial length scale ($L_{\text{slow}} = 2.0 L$), producing large, high-momentum shifts in the baseline wind vector.

\textbf{Fast Micro-Turbulence:} Modeled using a severely reduced length scale ($L_{\text{fast}} = 0.2 L$), generating erratic, high-frequency choppy gusts.

Both processes are driven by Gaussian white noise, with total aerodynamic variance distributed to maintain a balanced root-mean-square energy. 
This bimodal wind profile tests whether the cooperative policy learned true generalized disturbance rejection or merely over-fits the temporal frequencies of the training adversary.

\subsection{Training Instances}
The ablative baseline and HARL-AC approaches were tested using 10 random initialization each. 
Each individual run was executed for a total of 250 million environment steps each containing 254 intances to guarantee policy convergence across all curriculum stages and took about 6.5 hours of runtime on the hardware described previously.
The complete set of hyperparameters governing the neural network architectures and the Proximal Policy Optimization (PPO) updates are detailed in Appendix B.
The performance of these models was analyzed using four primary metrics: curriculum progression (retrospectively determined by a stage being progressed and never retreated to by a policy), success rate (percentage of successfully landed instances per episode) , crash rate (percentage of crashed instances per episode), and timeout rate (percentage of instances still flying after the 5 second time-limit per episode). 

\subsection{Evaluation Environment and Out-of-Distribution Testing} 
The evaluation setup for in distribution tests with both the DR and HARL-AC models is summarized in Table \ref{tab:evaluation_configs}. 
\begin{table}[!ht]
    \centering
    \caption{Evaluation Configurations.}
    \label{tab:evaluation_configs}
    \small
    {\setlength{\tabcolsep}{7pt}
    \renewcommand{\arraystretch}{1.25}
    \begin{tabular}{@{}llccc p{3.7cm}@{}}
        \toprule
        \textbf{Evaluation} & \textbf{Method} & \textbf{Runs} & \textbf{Trials/run} & \textbf{$\sigma$} & \textbf{Note} \\
        \midrule
        \multirow{2}{*}{Evaluation 1} & DR      & \multirow{2}{*}{10 each} & \multirow{2}{*}{10{,}000} & \multirow{2}{*}{0} & \multirow{2}{*}{\parbox[t]{3.7cm}{Only UAV and UR10 move; no wind.}} \\
        & HARL-AC & & & & \\
        \addlinespace[0.5em]
        \multirow{2}{*}{Evaluation 2} & DR      & \multirow{2}{*}{10 each} & \multirow{2}{*}{10{,}000} & \multirow{2}{*}{4} & \multirow{2}{*}{\parbox[t]{3.7cm}{Includes wind and coupled wave conditions; representative of sea state~4.}} \\
        & HARL-AC & & & & \\
        \addlinespace[0.5em]
        \multirow{2}{*}{Evaluation 3} & DR      & \multirow{2}{*}{10 each} & \multirow{2}{*}{10{,}000} & \multirow{2}{*}{7} & \multirow{2}{*}{\parbox[t]{3.7cm}{Increased wind and coupled wave conditions; representative of sea state~5.}} \\
        & HARL-AC & & & & \\
        \addlinespace[0.5em]
        \bottomrule
    \end{tabular}}
\end{table}

To gauge zero-shot robustness, the trained policies were evaluated in a out-of-distribution (OOD) testing environment designed to test spectral resilience. 
All DR and HARL-AC models were tested under higher sea states, represented through higher $\sigma$ values corresponding to sea states~7,~8, and~10.
Each of the 20 trials once again included 10,000 episodes to ensure the metrics better reflect zero-shot generalization without online adaptation to the unseen environments.

\section{Results and Discussion}
\subsection{Training results}

\begin{figure}[htbp]
    \centering
    \begin{subfigure}[b]{\textwidth}
        \centering
        \includegraphics[width=\textwidth]{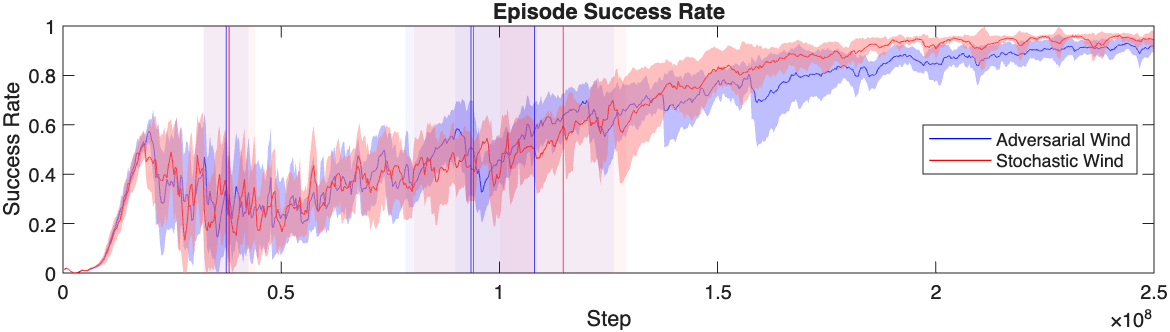}
        \caption{Average success rates. (higher is better)}
        \label{fig:episode_success}
    \end{subfigure}
    
    \vspace{1em}
    
    \begin{subfigure}[b]{\textwidth}
        \centering
        \includegraphics[width=\textwidth]{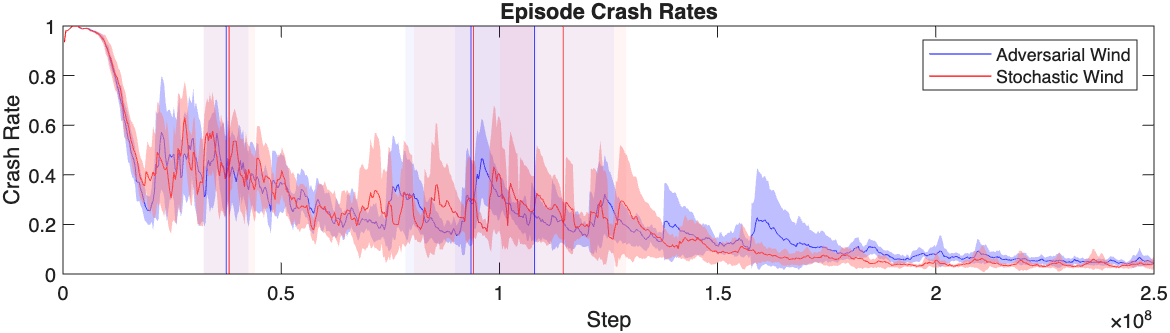}
        \caption{Average crash rates. (lower is better)}
        \label{fig:episode_crash}
    \end{subfigure}

    \vspace{1em}
    
    \begin{subfigure}[b]{\textwidth}
        \centering
        \includegraphics[width=\textwidth]{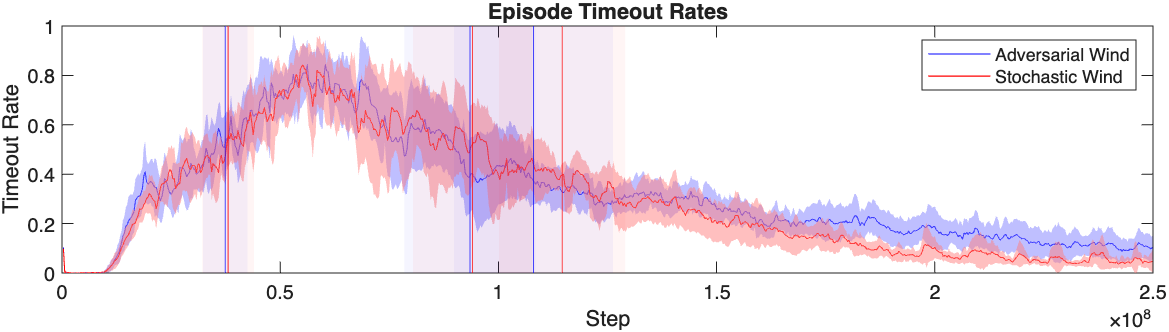}
        \caption{Average timeout rates. (lower is better)}
        \label{fig:episode_timeout}
    \end{subfigure}
    
    \caption{Averaged training metrics recorded over 10 repetitions during training of the HARL-AC (Adversarial) and DR (Stochastic) training paradigms. 
    The vertical bars indicate averaged transition points through the four stages of the curriculum. 
    Shaded regions depict the $95\%$ confidence interval around the mean results shown.}
    \label{fig:training_results}
\end{figure}

In Fig. \ref{fig:training_results}, the vertical solid bars and shaded rectangle around represent the mean and $95\%$ confidence interval (CI) around the stage transitions.
From left to right, these three bars denote the transitions from stage $0$ to $1$, $1$ to $2$, and $2$ to $3$, respectively. 
The line graphs display the performance metrics averaged across the 10 Monte Carlo trials conducted over 250 million training episodes; the shaded regions representing the corresponding $95\%$ CI.

The HARL-AC model converges marginally quicker in the earlier stages. 
Stage 0 to 1 and 1 to 2 transitions average at $37$ and $93$ million steps as opposed to the $38$ and $94$ million steps of the DR approach respectively.
However, the HARL-AC model demonstrates an improved ability to adapt to the increased difficulty of the later stages, evidenced by the greater discrepancy seen in the transition between stage 2 to 3 where the HARL-AC approach transitions at $108$ million steps as opposed to the $114$ million steps the DR approach required. 

During stage 3, the adversarial model exhibits spikes in crash rates as seen in Fig. \ref{fig:episode_crash} that correspond directly to dips in the success rate in Fig. \ref{fig:episode_success} at $138$ and $159$ million steps, though the success rate subsequently recovers. 
This behavior is expected, as it indicates the wind adversary (Team B) is continuously learning and discovering new strategies to cause the cooperative team (Team A) to fail. 
The HARL-AC model however does settle in these tests at a success rate of $90.8\%$, falling slightly short of the DR approach which settles around $94.3\%$. 
This is thought to stem from the added difficulty the adversarial wind agent poses as it can observe the effect of the actions chosen and strives to drive the total system back towards a $30\%$ win rate. 

The quicker overall convergence is hypothesized to be a result of the adversarial wind agent exposing the cooperative agents to a larger portion of the search space compared to the DR approach, ultimately allowing the cooperative team to derive a more generalized and robust solution.
To confirm this hypothesis an exploration into recordings of the state-ranges and wind force characteristics exposed to the cooperative team in each episode must be conducted.

Finally, the adversarial models present slightly higher timeout rates shown by the gap between the two plots in Fig. \ref{fig:episode_timeout}. 
An exploration into the deployed models and evaluation rounds revealed that this is due to a strict 5-second timeout limit set during training. 
Under the trialed sea state conditions, the HARL-AC model takes a few seconds longer to successfully land the drone compared to the DR model, completely accounting for the elevated timeout rates of the HARL-AC derived policies.
In the context of landing a drone after flight, this addition of two to three seconds in landing time has little overall impact on the deployability of the proposed solution. 
 
\subsection{Out-of-distribution evaluation} 
The box plots shown in Fig. \ref{fig:ood_results} illustrate the difference in performance of the HARL-AC and DR models across increasingly difficult sea states. 
The out-of-distribution testing environment utilizes the same stochastic wind setup that was used to train the baseline DR. 
An adjustment is made to the sigma value to produce an environment corresponding to sea states 7, 8 and 10. 
The OOD test environment allows us to validate the performance of the models in out-of-distribution performance and explore the generalizability and robustness of the two modeling approaches. 
In addition to the DR and HARL-AC OOD tests, an additional benchmark approach 'Baseline SS4' is included. 
The benchmark results are produced from a heterogeneous multi-agent training environment in which the adversary and curriculum are excluded, and the cooperative team is only exposed to conditions representative of sea state 4.

\begin{figure}[htbp]
    \centering
    \begin{subfigure}[b]{0.95\textwidth}
        \centering
        \includegraphics[width=\textwidth]{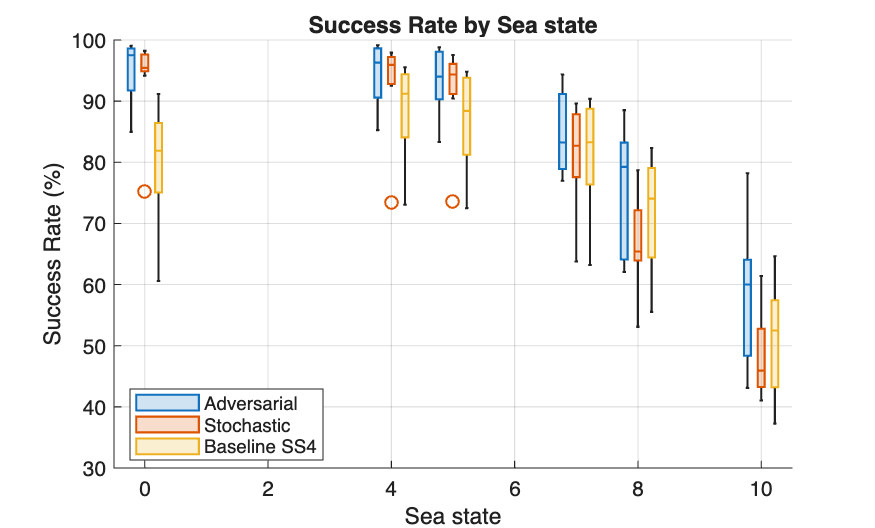}
        \caption{Success rates. (higher is better)}
        \label{fig:ood_success_rate}
    \end{subfigure}

    \vspace{1em}

    \begin{subfigure}[b]{0.95\textwidth}
        \centering
        \includegraphics[width=\textwidth]{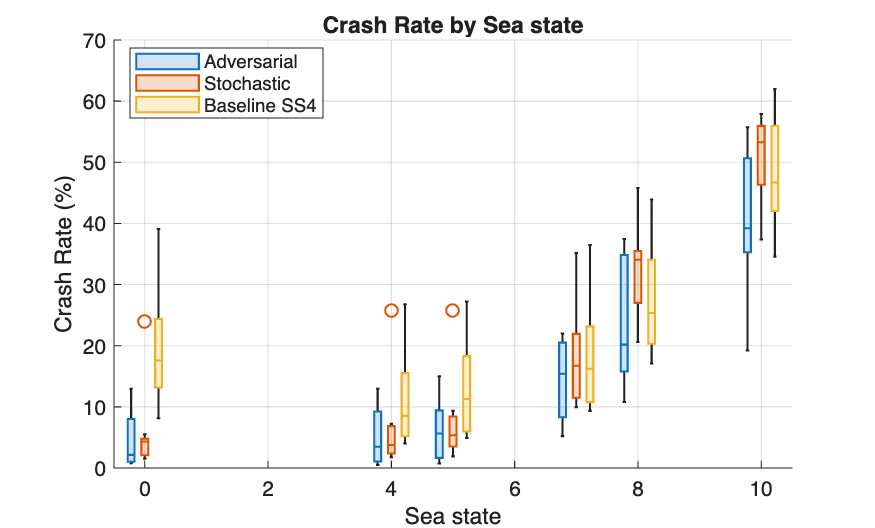}
        \caption{Crash rates. (lower is better)}
        \label{fig:ood_crash_rate}
    \end{subfigure}

    \caption{Average final evaluation metrics of the HARL-AC (Adversarial), DR (Stochastic) and Benchmark (Baseline SS4) approaches. 
    Results from sea states 0, 4, 5 corresponding to the the results shown from training stages shown in Fig. \ref{fig:training_results} as well as the out of distribution tests in sea states 7, 8, and 10 are included. 
    Boxes use a $95\%$ CI to determine outliers.}
    \label{fig:ood_results}
\end{figure}

Within the trained scenarios, sea state 0, 4, and 5, both the HARL-AC and DR models exhibits similar evaluation results.
The median performances are comparable with slightly greater variability in the results of the HARL-AC approach. Sea states 0, 4, and 5 see a success rate of $97.5\%$ vs $95.4\%$, $96.3\%$ vs $95.9\%$ and $94.0\%$ vs $94.3\%$ for the HARL-AC vs DR models respectively. 
This suggests the DR approach is performing as well as the HARL-AC models, but these results exclude the outlier indicated in Fig. \ref{fig:ood_success_rate} for the DR models. 
HARL-AC vastly outperforms the Baseline SS4 models, meanwhile the DR models seem to perform better at sea states 4 and 5 given the smaller box dimensions compared to Baseline SS4. 
Relying solely on these in-distribution environments to draw conclusions makes the strong assumption that the simulated training environment is perfectly representative of reality. 
During training the Baseline SS4 model achieved a success rate of $91\%$, however here the benchmark models fail to generalize in the simplified wind and waveless environment.
The sea state 0 median success rate of $81.9\%$ for the Baseline SS4 model is a clear example of an imperfect training environment representation. 

As the environment progresses into higher sea states unseen during training, all three models naturally experience a degradation in overall performance, however, the HARL-AC model demonstrates far greater robustness. 
As visible in Fig.~\ref{fig:ood_success_rate}, the adversarial model demonstrates a higher median success rate.
The HARL-AC model records median success rates of $83.2\%$, $79.3\%$, and $60\%$ for sea states 7, 8, and 10 whereas the corresponding DR success rates are lower at $82.6\%$, $65.4\%$, and $45.9\%$. 
This is also reflected in the lower mean crash rates where the performance metrics vary by a few percentage points up to sea state 7.
A notable increase in crash rate of $14\%$ is seen in both sea state 8 and 10 when comparing the DR models against the HARL-AC models.  
The Baseline SS4 models produce median results that fall between the HARL-AC and DR models with the caveat of even higher variability than the corresponding DR models.  

\begin{figure}[htbp]
    \ContinuedFloat
    \captionsetup{list=no}
    \centering

    \begin{subfigure}[b]{\textwidth}
        \centering
        \includegraphics[width=0.95\textwidth]{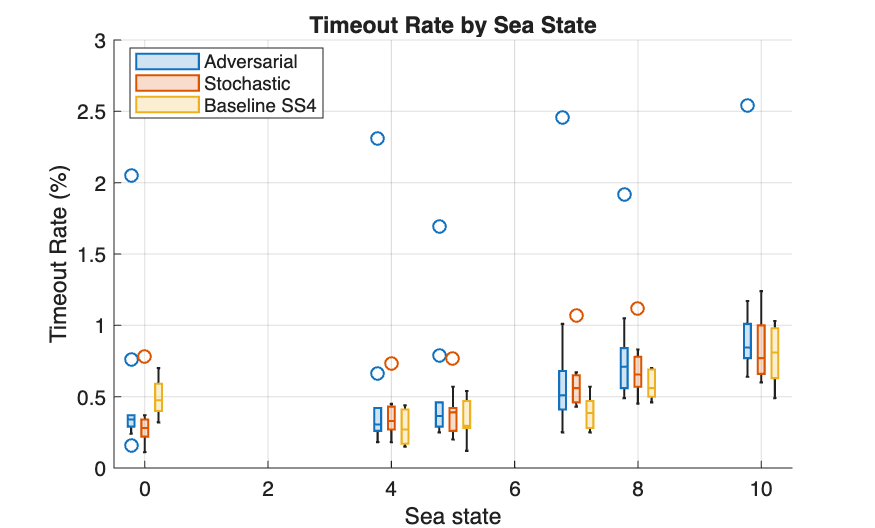}
        \caption{Timeout rates. (lower is better)}
        \label{fig:ood_timeout_rate}
    \end{subfigure}

    \caption{Average final evaluation metrics (continued).}
\end{figure}

Conversely, the timeout rate data shown in Fig.~\ref{fig:ood_timeout_rate} reveals a behavioral trade-off. 
Although timeout rates remain very low overall, under $3\%$, the HARL-AC model consistently yields slightly higher medians and presents several upper-bound outliers across the tested sea states. 
This is hypothesized to be an artifact of the learned policy - the adversarially trained agent has adopted a more wary, more cautious way of handling the turbulent conditions. 
By taking more time to safely navigate the stochastic disturbances, the HARL-AC model occasionally exceeds the strict time limits, as seen in fig. \ref{fig:ood_timeout_rate}, effectively trading a marginal increase in timeouts for a lower crash rate in OOD environments.
Empirical inspection of the HARL-AC model deployed to the OOD sea states showed the same result as previously, where the HARL-AC policy takes a few seconds longer.
This results in higher timeout rates but does not largely impact the operational deployability of the proposed model even into unseen environments.

Furthermore, direct empirical observations of the simulation trials reveal a distinct contrast in the control strategies developed by the HARL-AC and DR agents. 
The DR model exhibits highly reactionary behavior, it frequently over commits to fighting the current wind conditions, leaving it highly vulnerable to crashing when the wind unexpectedly changes in OOD sea states. 
This observation is supported by the higher relative crash out rates of the DR policies compared against the HARL-AC policies as seen in Fig. \ref{fig:ood_crash_rate}. 
In contrast, the HARL-AC model demonstrates far greater resistance to the unpredictable wind by actively taking its time. 
By adopting this wary and cautious approach, the adversarially trained agent successfully prioritizes overall stability over speed, explaining the increase in its timeout rates as a necessary trade-off to prevent crashes in unstable environments.

Ultimately, if the simulation environment is not perfectly representative of deployment conditions, the HARL-AC approach shows much greater resilience to this discrepancy.
The results shown illustrate that the HARL-C trained agent can generalize more effectively to unseen conditions and higher sea states. 
It is thus hypothesized that the adversarial model will be better equipped to handle unforeseen circumstances in live tests, providing a much smoother and safer real-world application.  

\section{Conclusion}
This study demonstrates that the proposed HARL-AC framework generates a more robust model compared to a Domain Randomization training paradigm for the challenging application of autonomous UAV landing in maritime settings. 
The increase in robustness of the HARL-AC agents in deployment and out-of-distribution tests is hypothesized to be a result of the adversarial wind agent exposing the cooperative UAV and robot-arm team to a much larger area of the search space during training.  
The dips and subsequent returns of the success rate corresponding to the rises and falls in crash rates in stage 3 of the curriculum learning suggests the adversarial agent continues to find new ways to challenge the cooperative team, even after reaching the final stage.
Consequently, while the training process is marginally faster overall, the continuous adversarial pressure forces the cooperative agents to continue learn more generalized recovery strategies.

The primary result of this training dynamic is that the adversarially trained agents develop a more cautious landing policy. 
While this cautious behavior results in slightly longer average landing times, it allows the HARL-AC model to generalize significantly better to unseen conditions and higher sea states.
Both the DR and Baseline SS4 models performed poorer by up to $16\%$ in success rates and $14\%$ on crash rate when compared to the HARL-AC model in OOD tests. 
The increased landing times do not meaningfully inhibit the deployability of the proposed methodology into a practical autonomous flight solution. 
If these simulated benefits extend to physical environments that inevitably violate the base assumptions that the simulation environment is perfectly accurate, we anticipate the HARL-AC model to be more robust against such environmental variances. 
Ultimately, by being exposed to greater intentional variability and continuous learning pressure, the HARL-AC policy prioritizes stability over speed. 
This adversarial framework provides a superior and more reliable landing solution for live applications of UAV landing on unstable platforms.

\bibliographystyle{splncs04}
\bibliography{references.bib}

\vspace{12pt}

\newpage
\appendix
\section{Adversarial wind pseudocode}
\begin{algorithm}[h]
    \begin{algorithmic}[1] 
        \Require Time step $\Delta t$, Curriculum stage $S$
        \Ensure Updated wind velocity $\vec{v}_{wind}$
        
        \If{$S \leq 1$} \Comment{Turn off wind for early stages}
            \State $\vec{\mu}_{wind} \gets 0$
            \State $\vec{x}_{turb} \gets 0$
            \State \Return $0$
        \EndIf
        
        \Statex
        \State \textbf{Step 1: Calculate smoothed adversarial wind}
        \State $\vec{w}_{adv} \gets \tanh(\text{action}) \times V_{max}$
        \State $\alpha_{ctrl} \gets \exp(-\Delta t / \tau_{mean})$ \Comment{Exponential moving average factor}
        \State $\vec{\mu}_{target} \gets \alpha_{ctrl} \vec{\mu}_{wind} + (1 - \alpha_{ctrl}) \vec{w}_{adv}$
        
        \State $\Delta \vec{w} \gets \vec{\mu}_{target} - \vec{\mu}_{wind}$
        \State $\Delta \vec{w} \gets \text{Clamp}(\Delta \vec{w}, -a_{max} \Delta t, a_{max} \Delta t)$ \Comment{Limit maximum acceleration}
        \State $\vec{\mu}_{wind} \gets \vec{\mu}_{wind} + \Delta \vec{w}$
        
        \Statex
        \State \textbf{Step 2: Add Dryden turbulence}
        \State $\alpha_{dryden} \gets (V \cdot \Delta t) / L$
        \State $N \gets \text{GenerateWhiteNoise}()$
        \State $\vec{x}_{turb} \gets (1 - \alpha_{dryden}) \vec{x}_{turb} + \sqrt{2 \alpha_{dryden}} \cdot \sigma \cdot N$
        
        \Statex 
        \State \textbf{Step 3: Output}
        \State $\vec{v}_{raw} \gets \vec{\mu}_{wind} + \vec{x}_{turb}$
        \State \Return $\vec{v}_{wind}$
    \end{algorithmic}
    \caption{Adversarial Wind Update }\label{alg:wind_update}
\end{algorithm}

\newpage
\section{HAPPO training hyperparameters}

\begin{table}[h!]

\centering

\caption{HAPPO Training Hyperparameters}

\label{tab:hyperparameters}

\small

\begin{tabular}{ll}

\toprule

\textbf{Parameter} & \textbf{Value} \\

\midrule

Total Environment Steps & $250,000,000$ \\

Episode Length (Steps) & $250$ \\

Network Hidden Layers (Actor \& Critic) & $[128, 128]$ \\

Activation Function & ReLU \\

Network Initialization & Orthogonal \\

Actor Learning Rate & $5 \times 10^{-4}$ \\

Critic Learning Rate & $5 \times 10^{-4}$ \\

Discount Factor ($\gamma$) & $0.99$ \\

GAE Parameter ($\lambda$) & $0.95$ \\

PPO Clip Range & $0.20$ \\

PPO Epochs & $5$ \\

Entropy Coefficient & $0.006$ \\

Max Gradient Norm & $10.0$ \\

Value Normalisation & True \\

Feature Normalisation & True \\

Parameter Sharing & False \\

\bottomrule

\end{tabular}

\end{table} 
\newpage

\section{Active reward terms}
\begin{table}[hb]
    \centering
    \caption{Active reward terms.}
    \label{tab:rewards}
    \small
    \begin{tabular}{llcl}
        \toprule
        \textbf{Term} & \textbf{Team} & \textbf{Weight} & \textbf{Active condition} \\
        \midrule
        \multicolumn{4}{l}{\textit{Team A (cooperative)}} \\
        \midrule
        \texttt{distance\_to\_goal} & A & 30.0 & always \\
        \texttt{proximity\_bonus} & A & 25.0 & $d < 0.25$~m \\
        \texttt{smooth\_landing} & A & 18.0 & $d < 0.25$, $\sum v_b^2 < 10$ \\
        \texttt{time\_shaping} & A & 0.5 & always \\
        \texttt{alignment\_reward} & A & 2.5 & $d < 0.90$~m \\
        \texttt{magnet\_reward} & A & 1000.0 & sustained capture \\
        \texttt{orientation} & A (arm) & 2.5 & always \\
        \texttt{arm\_go\_safe} & A (arm) & 0.1 & far + unsafe \\
        \texttt{arm\_hold\_still} & A (arm) & 0.08 & far + safe \\
        \texttt{arm\_near\_jitter} & A (arm) & 0.025 & near (penalty) \\
        \midrule
        \multicolumn{4}{l}{\textit{Team B (adversary)}} \\
        \midrule
        $-\,r_{\text{Team A}}$ & B & 1.0 & always (zero-sum) \\
        \texttt{wind\_distance} & B & 3.0 & always \\
        \texttt{wind\_instability} & B & 2.0 & always \\
        \texttt{wind\_misalignment} & B & 2.0 & always \\
        \texttt{wind\_crash\_bonus} & B & 25.0 & drone OOB \\
        \texttt{wind\_attrition} & B & 8.0 & timeout w/o capture \\
        \texttt{wind\_force\_penalty} & B & 0.1 & always (penalty) \\
        \bottomrule
    \end{tabular}%
\end{table}

\end{document}